\documentclass[10pt,twocolumn,letterpaper]{article}

\usepackage{iccv}
\usepackage{times}
\usepackage{epsfig}
\usepackage{graphicx}
\usepackage{amsmath}
\usepackage{amssymb}

\usepackage[pagebackref=true,breaklinks=true,letterpaper=true,colorlinks,bookmarks=false]{hyperref}

\iccvfinalcopy 

\def\iccvPaperID{10238} 

\ificcvfinal\fi

\begin{document}

\title{Deep Consensus Learning}

\author{Wei Sun and Tianfu Wu~\thanks{T.Wu is the corresponding author.}\\
Department of Electrical and Computer Engineering and the Visual Narrative Initiative\\
NC State University, Raleigh, NC 27695\\
{\tt\small \{wsun12, tianfu\_wu\}@ncsu.edu}}

\maketitle
\ificcvfinal\thispagestyle{empty}\fi

\begin{abstract}
   Both generative learning and discriminative learning have recently witnessed remarkable progress using Deep Neural Networks (DNNs). For structured input synthesis and structured output prediction problems (e.g., layout-to-image synthesis and image semantic segmentation respectively), they often are studied separately. This paper proposes deep consensus learning (DCL) for joint layout-to-image synthesis and weakly-supervised image semantic segmentation.
   The former is realized by a recently proposed LostGAN approach~\cite{LostGAN}, and the latter by introducing an inference network as the third player joining the two-player game of LostGAN. 
    Two deep consensus mappings are exploited to facilitate training the three networks end-to-end: Given an input layout (a list of object bounding boxes), the generator generates a mask (label map) and then use it to help synthesize an image. The inference network infers the mask for the synthesized image. Then, the latent consensus is measured between the mask generated by the generator and the one inferred by the inference network. For the real image corresponding to the input layout, its mask also is computed by the inference network, and then used by the generator to reconstruct the real image. Then, the data consensus is measured between the real image and its reconstructed image. The discriminator still plays the role of an adversary by computing the realness scores for a real image, its reconstructed image and a synthesized image. 
   In experiments, our DCL is tested in the COCO-Stuff dataset~\cite{caesar2018coco}. It obtains compelling layout-to-image synthesis results and weakly-supervised image semantic segmentation results. 
\end{abstract}


\section{Introduction}
\subsection{Motivations and Objectives}

\begin{figure}
    \centering
    \includegraphics[width=0.5\textwidth]{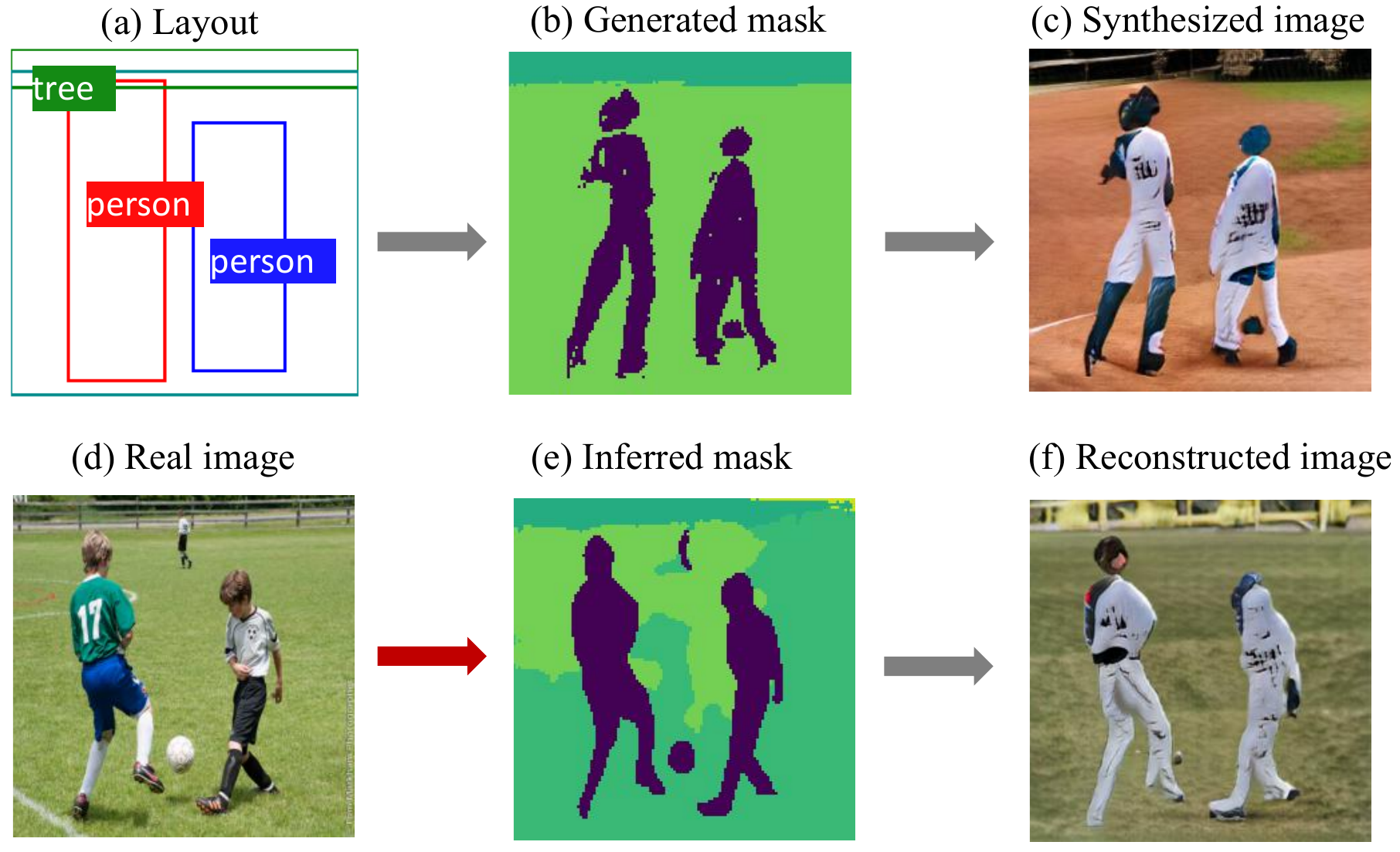}
    \caption{Illustration of the proposed deep consensus learning (DCL). 
    We consider joint layout-to-image synthesis (Top) and weakly-supervised image semantic segmentation (Bottom). The former is realized by a LostGAN~\cite{LostGAN,sun2020learning} pipeline which first generates a mask (b) from an input layout (a), and then synthesizes an image (c) conditioned on both (a) and (b). The latter is implemented by an inference network which is only supervised by the LostGAN in training without using ground-truth label maps (i.e., weakly-supervised). (b) and (c) form the synthetic training data for the inference network, which leads to \textbf{the latent consensus} mapping between (b) and the inferred label map for (c) by the inference network. Based on (a) and (e), the generator computes the reconstructed image (f). Then, (d) and (f) form \textbf{the data consensus mapping}.  See text for details.}
    \label{fig:DCLExample} 
\end{figure}

\begin{figure*}
    \centering
    \includegraphics[width=0.95\textwidth]{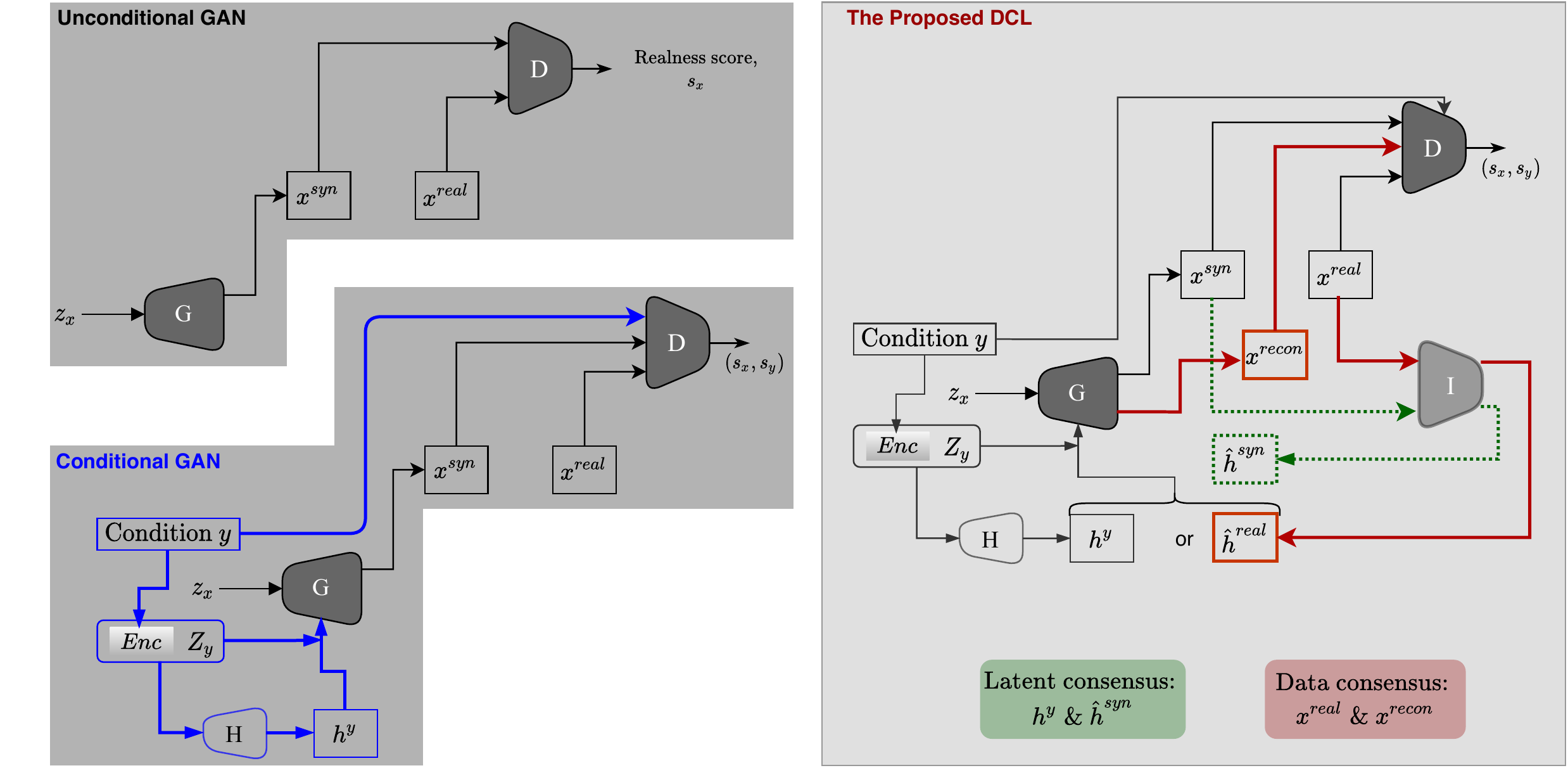}
    \caption{A systematic high-level overview of the proposed deep consensus learning (DCL) on top of unconditional and conditional GANs. 
     For layout-to-image synthesis, the condition $y$ is the spatial layout. A mask (semantic label map) $h^y$ is learned to facilitate better image synthesis to realize layout-to-mask-to-image synthesis in LostGANs~\cite{LostGAN,sun2020learning}.  The proposed DCL introduces an inference network $I$ to form a three-player game setting. The inference network is an image semantic segmentation network which is trained in a weakly-supervised way. The joint learning of the three networks are induced by two deep consensus mappings: \textbf{the data consensus} (arrows in red) between an real image $x^{real}$ and its reconstruction $x^{recon}$, and   \textbf{the latent consensus} (dotted arrows in green) between the semantic label map $h^{y}$ and the one $\hat{h}^{syn}$ inferred from the synthesized image $x^{syn}$. The former entails that the inference network can infer a meaningful label map $\hat{h}^{real}$ for a real image $x^{real}$. The latter entails the generator network to synthesize images $x^{syn}$ that are aligned with the conditions encoded by the semantic label map $h^y$. Best viewed in color.}
    \label{fig:DCL} 
\end{figure*}

Generative learning and discriminative learning are two of the most important aspects of machine (deep) learning. Both of them have witnessed remarkable progress with deep neural networks (DNNs) fueled by big data in the past decade. The goal of generative learning is to explain away the data itself in the spirit as Richard Freynman once famously said ``what I cannot create, I do not understand.". It is either to learn $p(x)$, the distribution of data $x$ such as images, which is often called \textit{unconditional data (image) synthesis}, or to learn $p(x|y)$, the conditional distribution of data (e.g., images) $x$ conditioned on $y$, which is usually called \textit{conditional data (image) synthesis}. $y$ represents class labels, bounding boxes, pixel-level label maps, or text descriptions, to name a few. For high-dimensional data such as images, generative learning that can generate high-fidelity data is an extremely challenging problem. On the contrary, the goal of discriminative learning is to perform prediction based on input data $x$, that is to learn $p(y|x)$. The prediction may not be induced from truly understanding the data and could take shortcut solutions~\cite{geirhos2020shortcut}. \textit{For structured input synthesis and structured output prediction problems}, the two lines of work have not been integrated to help each other from scratch in an end-to-end way. This paper takes one step forward by proposing \textbf{deep consensus learning} for joint layout-to-image synthesis and weakly-supervised image semantic segmentation (Fig.~\ref{fig:DCLExample} and Fig.~\ref{fig:DCL}).  

In computer vision, to enable flexible user control, conditional image synthesis, especially \textit{structured input image synthesis} such as the layout-to-image problem~\cite{zhao2020layout2image,LostGAN} (the top of Fig.~\ref{fig:DCLExample}), has emerged as an increasingly interesting yet challenging generative learning problem. The recent GauGANs~\cite{park2019semantic} showed that high-fidelity images can be generated based on input annotated label maps under an image-to-image translation framework. Concurrent to GauGANs, the layout-to-mask-to-image pipeline and the instance-sensitive layout-aware feature normalization (ISLA-Norm) scheme were proposed in the LostGANs~\cite{LostGAN,sun2020learning} for jointly generating the masks on-the-fly (e.g., (b) in Fig.~\ref{fig:DCLExample}). On the other hand, image semantic segmentation is a long-standing \textit{structured output prediction} problem (the bottom of Fig.~\ref{fig:DCLExample}), which requires time-consuming and often error-prone labeling efforts for collecting training data. Developing weakly-supervised image semantic segmentation that exploits only bounding box annotations in training is a promising direction. However, designing proper objective functions for weakly-supervised learning is a challenging problem, especially for structured output prediction tasks. \textit{Generative learning via explaining away the data can serve as the objective function for weakly-supervised discriminative learning}. However, it remains a challenging problem that how this bridge between generative learning and discriminative learning should be realized with stable end-to-end training. The proposed DCL (Fig.~\ref{fig:DCL}) attempts to pave a way towards addressing the challenge.

\subsection{Method Overview}
Fig.~\ref{fig:DCL} illustrates the proposed deep consensus learning (DCL), which has the following capabilities:

\textbf{Generative learning helps discriminative learning}. To facilitate learning layout-to-image synthesis, we adopt the layout-to-mask-to-image pipeline proposed in the LostGANs~\cite{LostGAN,sun2020learning}. First, the layout $y$ (i.e., the conditions in conditional generative learning) is encoded using a label embedding module $Enc(y)$. $Enc(y)$ is then concatenated with the style latent code $Z_y$. A light-weight mask generator $H$ is used to generate the initial label map $h^{y}$. The generator $G$ synthesizes an image $x^{syn}$ based on a sampled latent code $z_x$, $(Enc(y), Z_y)$, and $h^{y}$. The latter two are used to control the feature normalization modules in $G$ to realize conditional image synthesis with style and layout control. 
The LostGANs have shown promising results in layout-to-image synthesis with consistent masks $h^y$ and synthesized images $x^{syn}$ (see (b) and (c) in Fig.~\ref{fig:DCLExample}). We introduce an inference network for computing the semantic segmentation mask for an input image. For the inference network $I$, $(x^{syn}, h^{y})$ are used as ``supervised" training data, which leads to \textbf{the latent consensus} between $h^{y}$ and $\hat{h}^{syn}$ that inferred from $x^{syn}$. By doing so, \textit{the generator $G$ plays the role of a learnable objective function for the inference network $I$}. From a theoretical viewpoint, with the optimal generator $G$, the inference network $I$ can be trained to outperform the counterpart supervisedly trained on a given finite dataset with annotated semantic label maps, since the optimal $G$ is assumed to be capable of capturing the underlying distribution and of enabling extrapolation.  

\textbf{Discriminative learning helps generative learning}. In addition to the discriminator $D$, the inference network $I$ also provides feedback to the generator. Based on the inferred label map $\hat{h}^{real}$ for an real image $x^{real}$, the generator is induced to synthesize $x^{recon}$ that reconstructs $x^{real}$. This leads to \textbf{the data consensus} between $x^{real}$ and its reconstruction $x^{recon}$. Similarly, from a theoretical viewpoint, with the optimal inference network $I$ (i.e., an oracle image parser), the generator $G$ can be trained to generate high-fidelity images. This can be seen from the great results by GauGANs~\cite{park2019semantic}, which use annotated semantic label map in training and synthesis. 

\textbf{Joint generative and discriminative learning with deep consensus}. The three networks are trained end-to-end. The generator and discriminator networks form GANs~\cite{goodfellow2014generative} and realize layout-to-mask-to-image as done in LostGANs~\cite{LostGAN,sun2020learning}. The inference and generator networks cooperate similar in spirit to the variational auto-encoders (VAEs)~\cite{kingma2013auto,rezende2014stochastic} and the Kullback-Leibler divergence triangle learning framework~\cite{han2019divergence}. The inference network realizes weakly-supervised image semantic segmentation.

\section{Related Work}
\textbf{Generative Learning and Image Synthesis.}
Generative learning seeks probabilistic models that are capable of explaining away data under the maximum likelihood estimation (MLE) framework, which often entails introducing multiple layers of hidden variables, which in turn popularizes DNNs. There are two types of realizations in the literature: top-down generator models~\cite{goodfellow2014generative,han2017alternating,kingma2013auto,rezende2014stochastic,mnih2014neural,yu2017unsupervised,xie2019learning,xie2018cooperative} that are non-linear generalization of factor analysis~\cite{rubin1982algorithms}, and bottom-up energy-based models~\cite{lecun2006tutorial,ngiam2011learning,finn2016connection,kim2016deep,zhao2016energy,xie2016theory,nijkamp2019learning,kumar2019maximum,LazarowJT17} that are in the form of a Gibbs distribution or Boltzmann distribution. 
The proposed DCL is built on a top-down generator model.  

 Top-down generator models first sample a latent code $z\sim \mathcal{N}(0, 1_d)$, and then generate data using an explicit function, $x=f(z;\Theta)+\epsilon$. $f(;\Theta)$ is often parameterized by a DNN and $\epsilon$ is assumed to be Gaussian white noise with zero mean and variance $\sigma^2$ (i.e., $\epsilon\sim \mathcal{N}(0, \sigma^21_d)$). 
 To mitigate the difficulty of likelihood-based learning, variational autoencoders (VAEs)~\cite{kingma2013auto,rezende2014stochastic} introduce additional inference networks for computing $q(z|x)$ and work with a lower bound of the likelihood. GANs~\cite{goodfellow2014generative} bypass the modeling and learning of the distribution for $[z|x]$, and instead introduce a discriminator network to push the generator to generate high-fidelity data. Many variants of GANs have been proposed to improve the training stability~\cite{arjovsky2017wasserstein,zhao2016energy,karras2017progressive,kurach2018gan}. GANs are able to synthesize realistic and high resolution images under various settings, including both unconditional~\cite{arjovsky2017wasserstein, karras2017progressive,radford2015unsupervised,karras2018style} and conditional tasks~\cite{odena2017conditional,miyato2018cgans,brock2018large} including layout-to-image synthesis~\cite{johnson2018image,hong2018inferring,hinz2019generating,li2019object}. The proposed deep consensus learning is built on conditional GANs~\cite{miyato2018cgans}, more specifically on the LostGANs~\cite{LostGAN,sun2020learning}. But, the proposed method extends the LostGANs from conventional two-player settings to three-player settings, and integrates ideas of VAEs in a different way that the inference network does not directly compute the latent codes, but hidden states (label maps in the layout-to-image task). The proposed method also aims at addressing the problem of joint learning layout-to-image synthesis and weakly-supervised image semantic segmentation. 

\textbf{Discriminative Learning and Weakly-Supervised Image Semantic Segmentation.}
Discriminative learning seeks models that are capable of telling apart data in one class from data in another class among a predefined task output space, without explicitly entailing understanding the data itself. 
Image semantic segmentation, or image parsing, is a structured output prediction problem. Image parsing~\cite{tu2005image} has long been recognized as one of the grand challenges in computer vision. With the recent resurgence of DNNs, image parsing has achieved tremendous progress. 
Fully-supervised DNN-based image semantic segmentation methods often consist of two components: a convolutional neural network (CNN) backbone such as ResNets~\cite{he2016deep}, U-Net~\cite{ronneberger2015u}, Hourglass networks~\cite{newell2016stacked} and HRNets~\cite{sun2019high}, and a head classifier focusing on exploiting structured output information (i.e. context information) such as  FCN~\cite{long2015fully}, Atrous convolutions in the DeepLab methods~\cite{chen2017rethinking,deeplabv3plus2018}, PSPNet~\cite{zhao2017pspnet} and OCRNet~\cite{YuanCW20}. In our proposed DCL, the inference network adopts a simple U-Net architecture under the vanilla image-to-map FCN setting. Fully-supervised image semantic segmentation methods assume to provide the ``upper-bound" in performance for weakly-supervised methods. Weakly-supervised image semantic segmentation methods can leverage image-level labels~\cite{ahn2019weakly,briq2018convolutional}, object bounding boxes~\cite{SongHOW19,HuDHDG18,ZhaoLW18}, and points~\cite{bearman2016s}, etc. Typically, class activation maps~\cite{ZhouKLOT16} are exploited to create pseudo ground-truth label maps. Most of existing weakly-supervised image semantic segmentation methods are tested in relatively smaller PASCAL VOC12 dataset~\cite{EveringhamGWWZ10}. Our proposed DCL adopts bounding boxes as weakly-supervised signals for image semantic segmentation. In our DCL, the inference network is trained from scratch without exploiting the ImageNet-pretrained CNN backbones and tested in the COCO-Stuff dataset~\cite{caesar2018coco}. 

\textbf{Joint Generative and Discriminative Learning.}
Although discriminator networks are used in GANs, they usually do not perform sophisticated discriminative learning tasks like multi-class classification, object detection and image semantic segmentation, and are discarded after training. One of very exciting development is the introspective neural network (INNs)~\cite{jin2017introspective,LazarowJT17}, which builds a single model which is simultaneously generative and discriminative, but entails a sequence of CNN classifiers for generative learning. Motivated by the WGAN~\cite{arjovsky2017wasserstein}, Wasserstein INNs~\cite{LeeX0T18} are further proposed to learn a single CNN model for both generative and discriminative modeling with promising results on unconditional image synthesis and image classification. The energy-based generative aspects of discriminative CNN classifiers also have been studied in~\cite{GrathwohlWJD0S20,xie2016theory}. In~\cite{zheng2019joint}, a joint learning framework of person re-id learning and image generation is proposed. In our proposed DCL, we integrate generative learning and discriminative learning using different CNNs for a joint structured input synthesis and structured output prediction problem. 

\textbf{Cycle-Consistency Learning and Dual Learning.} 
With the remarkable generation capabilities achieved by the recent (conditional) generative learning models, cycle-consistency has been studied as a powerful loss term in learning mappings between two different domains such as the image-to-image translation by CycleGANs~\cite{zhu2017unpaired} and DualGANs~\cite{YiZTG17}, which have inspired many different variants with great successes in different application domains. Our proposed DCL enjoys the spirit of cycle-consistency learning. Unlike CycleGANs that learn mappings between two different domains with ground-truth inputs and two different generator networks, our DCL exploits latent and implicit cycle-consistency within the same generator and between the generator network and inference network. 

\textbf{Kullback-Leibler Divergence Triangle Learning.} More recently, the KL divergence triangle learning framework~\cite{han2019divergence,0001NZPZW20} is proposed to unify adversarial learning, variational learning and cooperative learning between three networks: generator, energy-based and inference networks. It develops a common objective function that is a symmetric and anti-symmetric combination of three KL divergences between the three models. Motivated by the divergence triangle learning framework, our proposed DCL also aims at integrating three networks, but is formulated under a different way (Eq.~\ref{eq:objective}) and shows the capabilities of joint layout-to-image synthesis and weakly-supervised image semantic segmentation. 

\textbf{Our Contributions.} The proposed DCL makes three main contributions to the field of joint generative and discriminative learning: (i) Our DCL shows a steady pace of extending GANs and conditional GANs to the three-player settings. And, to our knowledge, our DCL is the first work integrating conditinoal GANs for weakly-supervised semantic segmentation. (ii) Our DCL provides an alternative method of applying latent and implicit cycle-consistency and duality for weakly-supervised structured output prediction tasks. (iii) Our DCL obtains compelling layout-to-image synthesis results and  weakly-supervised image semantic segmentation results in the COCO-Stuff dataset.  

\section{The Proposed Method}
\subsection{Problem Formulation}
\textbf{Layout-to-Image.} Denote by $\Lambda$ an image lattice (e.g., $256\times 256$) and by $x$ an image defined on the lattice. Let $y=\{(\ell_i, B_i)_{i=1}^m\}$ be a layout consisting of $m$ labeled bounding boxes, where label $\ell_i\in \mathcal{C}$ (\eg, $|\mathcal{C}|=171$ in the COCO-Stuff dataset~\cite{caesar2018coco}), and a bounding box $B_i\subseteq \Lambda$. Different bounding boxes may overlap. Let $z_{x}$ be the latent code of image style and $z_{obj_i}$ the latent code of object instance style for $(\ell_i, B_i)$. Typically, the latent codes are randomly sampled from the standard Normal distribution, $\mathcal{N}(0,1)$ under the IID setting. Denote by $Z_y=\{z_{obj_i}\}_{i=1}^m$ the set of object instance style latent codes. The goal of layout-to-image synthesis is to learn a generator network, 
\begin{equation}
    x^{syn} = G(z_{x}, Z_{y}, y; \Theta_G), 
\end{equation}
where $\Theta_G$ collects all the parameters of the generator network. For conditional image synthesis, the style augmented conditions $(y, Z_y)$ are not directly used as input to the generator. Instead, they are used to control the affine transformation component in the feature normalization modules (such as BatchNorm~\cite{ioffe2015batch}) in the generator in different ways~\cite{de2017modulating,dumoulin2017learned,brock2018large,park2019semantic,park2019semantic,LostGAN}. 

\textbf{Layout-to-Mask-to-Image.} This pipeline is proposed in the LostGANs~\cite{LostGAN}, which generates a label map $h^y$ (i.e., imputing the hidden variables) for an input layout to facilitate image synthesis with better style and layout control,
\begin{equation}
h^y=H(Z_{y}, y;\theta_h),    
\end{equation} 
where $\theta_h$ collects all the parameters of the mask generator. Then, the generator network is rewritten as, 
\begin{equation}
    x^{syn} = G(z_{x}, h^y; \Theta_G), 
\end{equation}
where $h^y$ is used to compute instance-sensitve layout-aware affine transformation in the feature normalization modules in the generator~\cite{LostGAN,sun2020learning,park2019semantic}. 

\textbf{Image Semantic Segmentation.} Consider a real image $x^{real}$, let $\hat{h}^{real}$ be its label map inferred by an inference network $I$, we have, 
\begin{equation}
    \hat{h}^{real} = I(x^{real};\Theta_I), 
\end{equation}
where $\Theta_I$ collects all parameters of the inference network. Then, we can reconstruct the image $x^{real}$ using the generator network, 
\begin{equation}
    x^{recon} = G(z_{x}, \hat{h}^{real}; \Theta_G). 
\end{equation}

Similarly, we can obtain the inferred label map for a synthesized image $x^{syn}$, 
\begin{equation}
    \hat{h}^{syn} = I(x^{syn};\Theta_I).
\end{equation}

\textbf{Deep Consensus Learning.} The generator network and the inference network work cooperatively with two types of deep consensus: 
\begin{align}
    \text{Data consensus: } & x^{real}\xrightarrow{I} \hat{h}^{real}\xrightarrow[y, Z_y, z_x]{G} x^{recon} , \\
    \text{Latent consensus: } & y\xrightarrow[Z_y]{H}h^y\xrightarrow[y,Z_y, z_x]{G} x^{syn}\xrightarrow{I} \hat{h}^{syn} .
\end{align}
To reach the data consensus, the inference network is induced to infer a meaningful label map for a real image, which leads to the weakly-supervised image semantic segmentation capability. To reach the latent consensus, the generator network is induced to serve as a learnable objective function for the inference network, and by itself needs to generate high-fidelity images aligned with the label map ($h^y$ or $\hat{h}^{real}$). 
In training, a discriminator network $D(x, y;\Theta_D)$ is used to compute the realness score for an image $x\in\{x^{real}, x^{syn}, x^{recon}\}$, playing the role of an adversary. The output of $D$ consists of a list of realness scores for an input image and the corresponding $m$ objects defined by the bounding boxes $B_i$'s, 
\begin{equation}
D(x, y;\Theta_D) = (p_{img}, p_{obj_1},\cdots, p_{obj_m}). \label{eq:D}
\end{equation}

The data and latent consensus mappings enable end-to-end training of the three networks, $G(\cdot;\Theta_G), D(\cdot;\Theta_D)$ and $I(\cdot;\Theta_I)$ and the label map generator $H(\cdot;\theta_h)$. 

\subsection{Objective Function}
The objective function used in training is defined by,
\begin{alignat}{2}
 & \min_{\Theta_G, \theta_h, \Theta_I} \max_{\Theta_D} && \quad  \mathcal{L}(\Theta_G, \theta_h, \Theta_I, \Theta_D)= \label{eq:objective}\\  
  \nonumber & \mathbb{E}_{\substack{(x^{real}, y)\sim P,\\Z\sim \mathcal{N}}} \Bigg[&& ||x^{real}-x^{recon}||_1 +  ||F(x^{real})-F(x^{recon})||_1 +   \\ 
  \nonumber  & &&   ||F(x^{real})-F(x^{syn})||_1 \Bigg] + \\
  \nonumber & \mathbb{E}_{\substack{ y\sim P,\, Z\sim \mathcal{N}}} && \bigg[mKL(h^y||\hat{h}^{syn}) \bigg] +  \\
  \nonumber & \mathbb{E}_{\substack{(x^{real},y)\sim P, \\Z\sim \mathcal{N}}} \Bigg[&& l_D(x^{real}, y) + l_D(x^{recon}, y) + l_D(x^{syn}, y) \Bigg],  
\end{alignat}
where $P$ represents the data distribution of $(x^{real}, y)$, and $\mathcal{N}$ the standard Normal distribution, and $Z=(z_x, Z_y)$. 

\textit{The first term defines the data consensus}, consisting of the reconstruction loss between a real image $x^{real}$ and its reconstructed image $x^{recon}$, and the perceptual loss terms~\cite{johnson2016perceptual} which measure the $L1$ difference between features, $F(\cdot)$ extracted by an ImageNet~\cite{deng2009imagenet} classification pretrained network such as the VGG network~\cite{VGG}. 

\textit{The second term defines the latent consensus} which uses the mean Kullback-Leibler divergence between the two learned label maps across all pixels, 
\begin{equation}
    mKL(h^y||\hat{h}^{syn}) = \frac{1}{|\Lambda|} \sum_{\mathbf{x}\in \Lambda} KL\Big(h^y(\mathbf{x}) || \hat{h}^{syn}(\mathbf{x})\Big), 
\end{equation}
where both $h^y(\mathbf{x})$ and $\hat{h}^{syn}(\mathbf{x})$ represent the label probability vector of labels in the label-set $\mathcal{C}$. $h^y$ is the updated label map by the Generator (see Fig.~\ref{fig:generator}). Note that we use the KL divergence to preserve the intrinsic uncertainty in $h^y$ when it is treated as the ``ground-truth" label map in training the inference network on $x^{syn}$. We do not take {\tt argmax} of $h^{y}(\mathbf{x})$ among different categories at a pixel to form one-hot label vectors. By doing this, it is similar in spirit to the network distillation method~\cite{hinton2015distilling}.

\textit{The third term defines the adversary loss} computed by the discrminator network. $l_D(x, y)$ is defined by the hinge version~\cite{HingeLoss1,HingeLoss2} of the standard adversarial loss~\cite{goodfellow2014generative}, 
\begin{equation}
    l_D^{t}(x, y) = \begin{cases} \max (0, 1 + p_t);\, & \text{if } x = x^{syn} \text{ or } x^{recon} \\ \max (0, 1 - p_t);\, & \text{if } x=x^{real} 
     \end{cases}
\end{equation}
where  $t\in \{img, obj_1, \cdots, obj_m\}$ and $x\in\{x^{real}, x^{syn}, x^{recon}\}$. In the hinge loss, no penalty will occur if the score of a real image (or a real object instance) is greater than or equal to $1$, and the score of a fake image (or a fake object instance) is less than or equal to $-1$. The hinge loss is more aggressive than the real \emph{vs} fake binary classification in the vanilla GANs, which induces the discriminator to be more like an energy-based model~\cite{lecun2006tutorial,xie2016theory}.

\subsection{Implementation Details}\label{sec:implementation}\vspace{-2mm}
We adopt the settings of the generator and discriminator networks used in the LostGANs~\cite{LostGAN,sun2020learning}. To be self-contained, we first give a brief overview of LostGANs. We also propose a new implementation of updating $h^y$ or $\hat{h}^{real}$ using the intermediate features of the generator network. For the inference network, we use the U-Net~\cite{ronneberger2015u} for its simplicity and popularity in segmentation tasks.

\textbf{The layout encoder and the label map generator network $H(Z_{y}, y;\theta_h)$}. We encode the labels of bounding boxes in a layout. We use one-hot label vector for the $m$ object instances in a layout $y$, which results in a one-hot label matrix, denoted by $Y$, of the size $m\times d_{\ell}$, where $d_{\ell}=|\mathcal{C}|$ is the number of object categories. Label embedding is to learn a $d_{\ell}\times d_e$ embedding matrix, denoted by $W^{y}$. We have $\mathbb{Y} = Y \cdot W^y$,
where $\mathbb{Y}$ is a $m\times d_e$ matrix and $d_e$ represents the embedding dimension (e.g., $d_e=128$ in our experiments). 
Note that we do not exploit the geometric information (e.g., relative location and scale) of bounding boxes for simplicity, which may be 
potentially useful. $Z_{y}$ is concatenated with $\mathbb{Y}$, which represents the latent code matrix of the size $m\times d_{y}$ for controlling object instance styles (where $d_{y}=128$ in our experiments). With the concatenated $(\mathbb{Y}, Z_{y})$ as input, the (hidden) label map generator network computes the initial label map $h^y$. 

\begin{figure}[!h]
    \centering
    \includegraphics[width=0.48\textwidth]{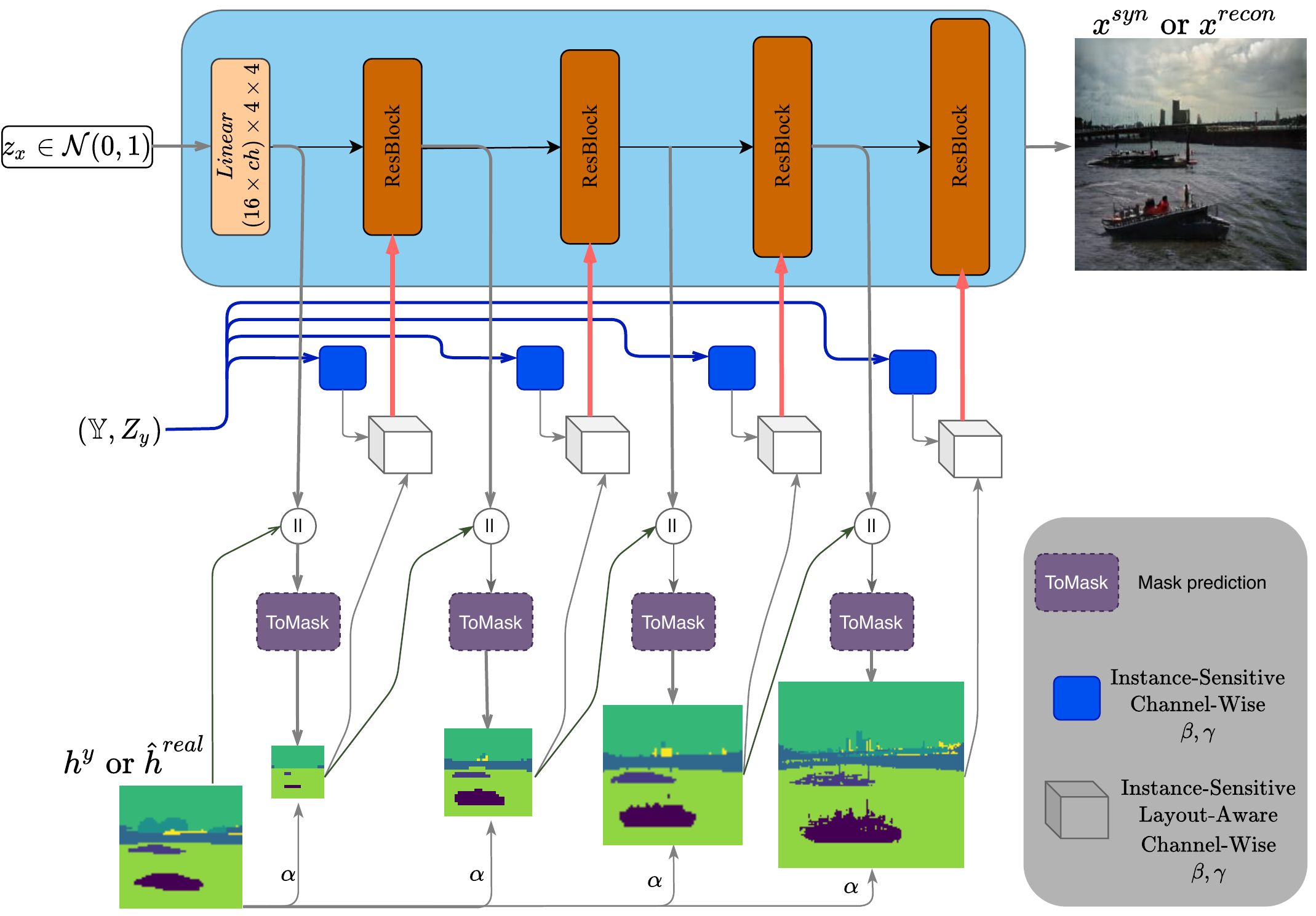}
    \caption{Illustration of the generator network and the Instance-Sensitive Layout-Aware feature Normalization scheme (ISLA-Norm). They are adopted and modified from the LostGANs~\cite{LostGAN} to be self-contained. See text for detail. }
    \label{fig:generator} 
\end{figure}
\textbf{The generator network and the ISLA-Norm.} Fig.~\ref{fig:generator} illustrates the network architecture and the ISLA-Norm, which are modified from the LostGANs~\cite{LostGAN,sun2020learning}. The generator network $G$ takes the image-level latent code $z_x$ as input and then transforms $z_x$ into a a feature map of spatial size $4\times 4$ with $16\times ch$ channels (e.g., $ch=16$ in our experiments) using  a linear fully-connected (FC) layer. A number of residual building blocks (ResBlocks)~\cite{he2016deep} is followed depending on the target resolution of image synthesis. Each ResBlock increases the resolution by a factor of $2$. The encoded layout $\mathbb{Y}$, the style latent code $Z_y$, and the generated label map $h^y$ (or the inferred one $\hat{h}^{real}$) are used to control the affine transformation parameters of feature normalization modules in ResBlocks, i.e., shifting parameters $\beta$ and rescaling parameters $\gamma$, to achieve instance-sensitive layout-aware image synthesis results.

\textbf{The ISLA-Norm.} A feature normalization module consists of two components: feature standardization and affine transformation. Let $f$ be an input 4-D feature map to the feature normalization module. $f_{b,c,h,w}$ is the feature response at a position $(b, c, h, w)$ using the convention index order of axes for batch, channel, and spatial height and width. In feature standardization, ISLA-Norm is the same as the BatchNorm~\cite{ioffe2015batch}, $\bar{f}_{b,c,h,w} = \frac{f_{b,c,h,w}-\mu_c}{\sigma_c}$, 
where the channel-wise mean $\mu_c$ and standard deviation $\sigma_c$ are pooled across the spatial dimensions from the entire batch.  In affine transformation, unlike the BatchNorm which learns channel-wise $\beta_c$ and $\gamma_c$ shared by all positions and all instances in a batch,  ISLA-Norm learns instance-sensitive layout-aware affine transformation parameters in two steps. Without loss of generality, let the batch size be $1$.  

\textit{Learning object instance-sensitive channel-wise $\beta_{m,c}$ and $\gamma_{m,c}$} based on  $(\mathbb{Y}, Z_y)$ computed for an input layout, as illustrated by the blue squares in Fig.~\ref{fig:generator}. Simple linear transformation functions are learned. Let $S=(\mathbb{Y}, Z_y)$, which is of the size $m\times (d_e+d_y)$. A linear transformation matrix $W^{S}$ of the size $(d_e+d_y)\times 2C$ is learned to compute $(\beta_{m,c}, \gamma_{m,c})=S\cdot W^{S}$, where $C$ is the number of channels.

\begin{figure*} [t]
	\centering
	\includegraphics[width=0.99\textwidth]{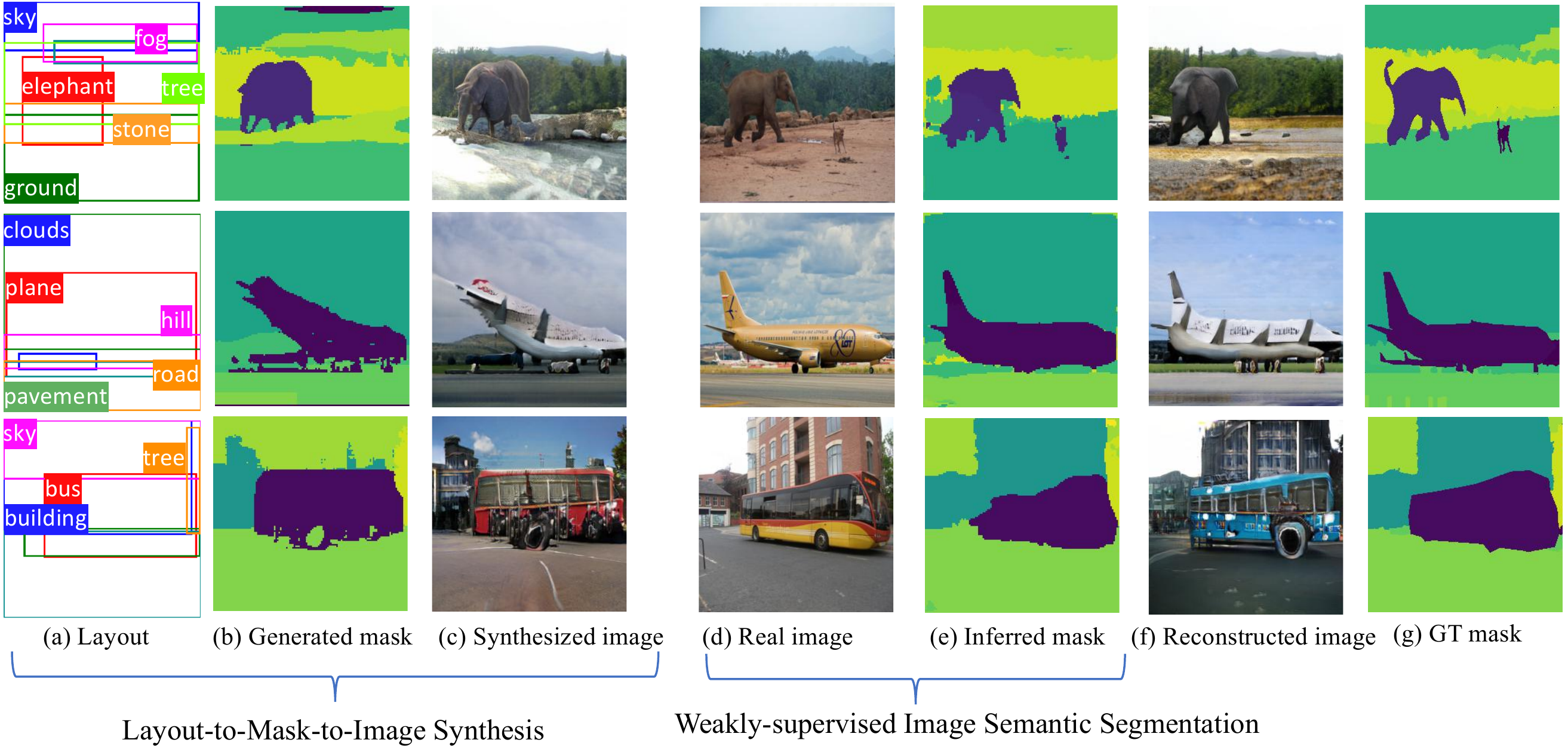}
	\caption{Examples of results by the proposed DCL for joint  layout-to-image synthesis and weakly-supervised image semantic segmentation in the COCO-Stuff dataset. 
	} 
	\label{fig:teaser}
\end{figure*}

\textit{Learning object instance-sensitive layout-aware $\beta_{c,h,w}$ and $\gamma_{c,h,w}$} by assembling $\beta_{m,c}$ and $\gamma_{m,c}$ based on the learned and iteratively updated semantic label map $h$ ($h$ starts with $h^y$ or $\hat{h}^{real}$). We use a different approach in updating the label map to better exploit intermediate features in the generator. We first concatenate the input feature map $f$ to a ResBlock (i.e., the output from the previous layer) and the input label map $h$ (resized to match the size of the feature map). Then, a light-weight ``ToMask" module is used to predict a new label map. In the meanwhile, the initial label map $h^y$ (or $\hat{h}^{real}$) is added to the new label map using a skip connection with a learnable weight $\alpha$. Based on the updated label map $h$, $\beta_{m,c}$ and $\gamma_{m,c}$ are unsqueezed into the corresponding bounding box $B_i$ weighted by the learned label probability in $h$ ($i=1, \cdots, m$) to form $\beta_{c,h,w}$ and $\gamma_{c,h,w}$, as illustrated by the grey cubes in Fig.~\ref{fig:generator}. Then, the feature response $\bar{f}_{b,c,h,w}$ is recalibrated by,
\begin{equation}
    \Tilde{f}_{b,c,h,w}=\gamma_{b,c,h,w}\cdot \bar{f}_{b,c,h,w} + \beta_{b,c,h,w},
\end{equation}
which helps control the layout and style of generated images at both image and object instance levels with reasonably high fidelity.

\textbf{The discriminator network.} Following LostGANs~\cite{LostGAN,sun2020learning}, it consists of a ResNet feature backbone, an image-level head classifier to compute $p_{img}$ (Eqn.~\ref{eq:D}), and an object-level head classifier to compute $(p_{obj_1, \cdots, p_{obj_m}})$ (Eqn.~\ref{eq:D}) using the bounding boxes $B_i$'s in the input layout as the region-of-interest (RoI), as done in the Faster-RCNN object detection systems~\cite{FasterRCNN}.

\section{Experiments}
We test our DCL in the COCO-Stuff dataset~\cite{caesar2018coco}. We evaluate our DCL at the resolution of 256$\times$256. We built on the publicly available PyTorch source code of LostGANs~\footnote{\url{https://github.com/iVMCL/LostGANs}}. \textbf{Our source code will be released too.}

\textbf{Training details.}
Following the settings in the LostGANs-V2~\cite{sun2020learning}, we use the Adam optimizer~\cite{kingma2014adam} with ($0$, $0.999$) as ($\beta_1$, $\beta_2$) for both the generator network and the discriminator network, and with ($0.9$, $0.999$) for the inference network. The learning rate is set to $0.0001$ for all the three networks. Synchronized BatchNorm~\cite{SyncBN} with feature statistic accumulated over all devices is utilized in the generator network. Batch size is set to $48$ on $4$ Tesla V100 GPUs.
FP16 mixed precision training with the NVIDIA Apex library~\footnote{\url{https://github.com/NVIDIA/apex}} is used to speed up the training. We report the results based on the model trained with 25 epochs due to some instability issues of the three-player optimization problem.

\subsection{Layout-to-image synthesis results}
\textbf{Baselines.} We compare with the LostGAN-V2~\cite{sun2020learning} and the Grid2Im method~\cite{ashual2019specifying}.   

\textbf{Evaluation Metrics.} It still is an open problem of automatically evaluating image synthesis results in general. For the layout-to-image synthesis, we adopt three widely used metrics: the \textit{Inception Score} (IS)~\cite{salimans2016improved}, the \textit{Fr\`echet Inception Distance} (FID)~\cite{FID}, and the \textit{Diversity Score} (DS) using the LPIPS method~\cite{zhang2018unreasonable}. 

\begin{table}[!ht]
    \centering
    \small{
    \resizebox{0.48\textwidth}{!}{
    \begin{tabular}{c|c|c|c|c}
    \hline 
     Model & \#Epoch  & IS$\uparrow$ & FID$\downarrow$ & DS$\uparrow$ \\ \hline
        Real Images & - & 28.10 $\pm$ 1.60 & - & - \\ \hline
        Grid2Im~\cite{ashual2019specifying} (GT Layout) & - &15.22  $\pm$ 0.11 &  65.95 & 0.34 $\pm$ 0.13 \\ \hline 
        LostGAN-V2~\cite{sun2020learning} & 100 & 18.01 $\pm$ 0.50 & 42.55 & 0.55 $\pm$ 0.09 \\ \hline
        Our DCL & 25 & 15.40 $\pm$ 0.25 & 50.45 & 0.51 $\pm$ 0.10 
        \\ \hline
    \end{tabular}}}
    \\ [1ex]
    \caption{Quantitative comparisons using the Inception Score (IS, higher is better), FID (lower is better) and Diversity Score (DS, higher is better) evaluation metrics in the COCO-Stuff~\cite{caesar2018coco}.}
    \label{tab:gen_evaluation} 
\end{table}

\begin{figure}[!ht]
    \centering
    \includegraphics[width=1.0\linewidth]{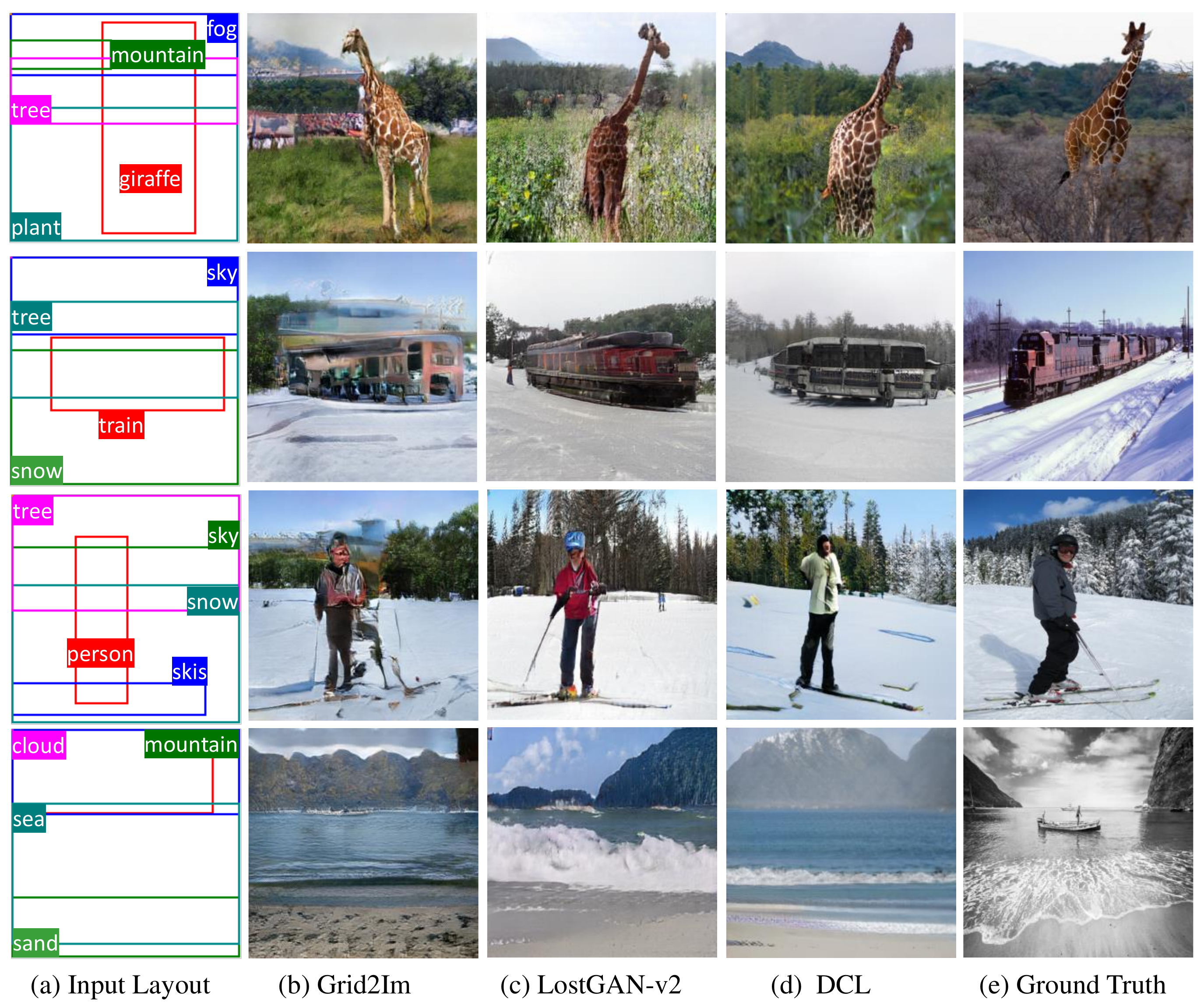}
    \caption{Generated samples from given layouts in COCO-Stuff by different models. (a) Input Layout, and Images generated by (b) Layout2Im~\cite{zhao2020layout2image} $64\times 64$, (c) Grid2Im~\cite{ashual2019specifying}  $256\times 256$, (d) LostGAN-V2~\cite{sun2020learning} $256\times 256$, (e) our DCL $256\times 256$, and (f) Ground Truth.}
    \label{fig:generator_evaluation} 
\end{figure}

\textbf{Results.} Table~\ref{tab:gen_evaluation} shows the comparisons. Although our DCL is only trained with 25 epochs, it obtains better performance than the Grid2Im method, and is not significantly worse than the LostGAN-V2 method. Fig.~\ref{fig:teaser} and  Fig.~\ref{fig:generator_evaluation} show examples of different methods. We believe that our DCL will obtain performance on a par with, or even better than, the LostGAN-V2 if the instability issues are resolved and more epochs are executed in training, since our DCL is built on the LostGAN-V2.   

\begin{figure*}[t]
    \centering
    \includegraphics[width=0.99\linewidth]{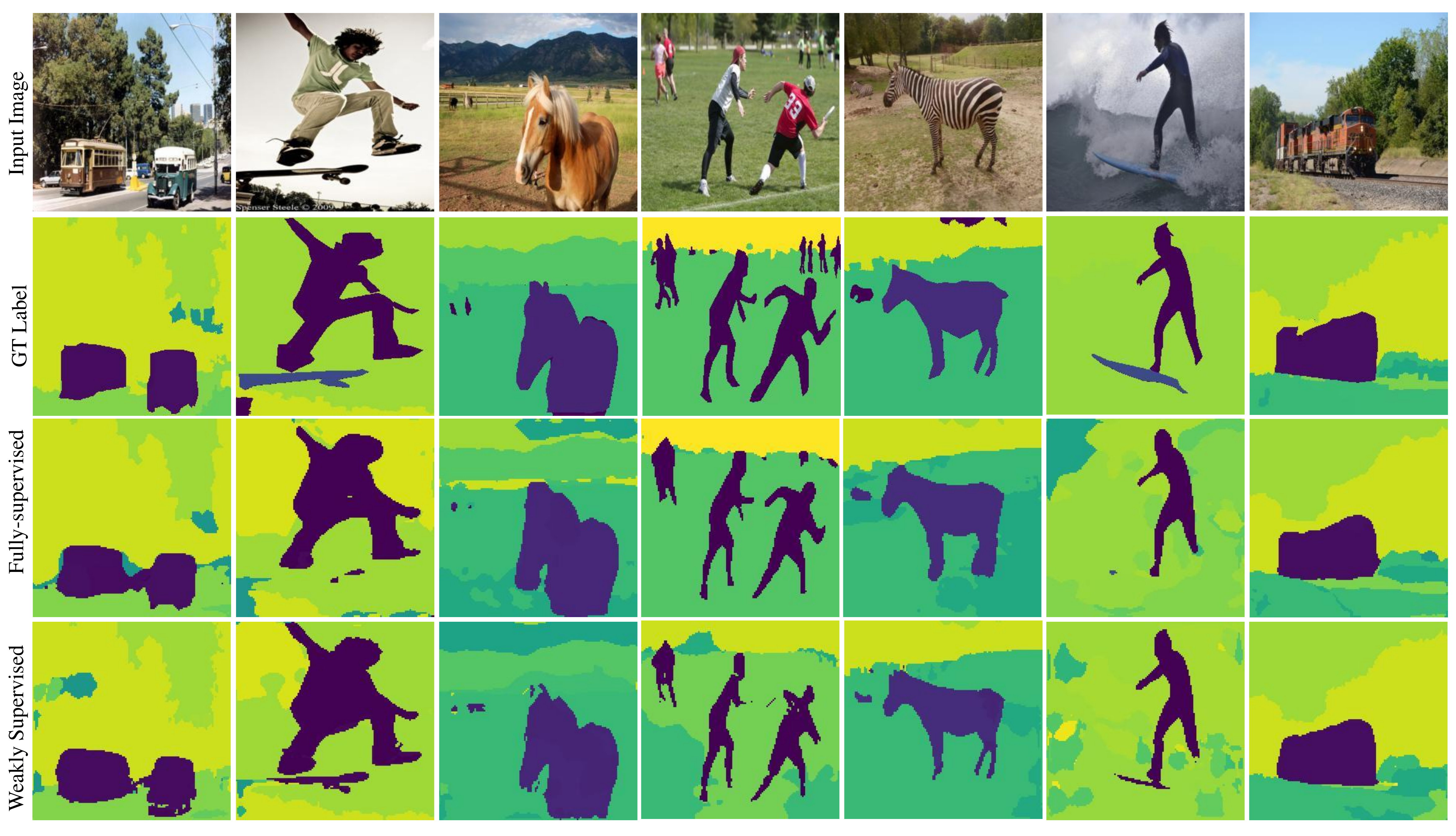}
    \caption{Examples of semantic segmentation results by the weakly-supervised inference network in our DCL and its fully-supervised counterpart.}
    \label{fig:segmentation_evaluation} 
\end{figure*}

\begin{table*}
    \centering
    \small{
    \resizebox{0.7\textwidth}{!}{
    \begin{tabular}{c|c|c|c|c|c}
    \hline
    \textbf{Methods} & \#Epoch & \textbf{Class Accuracy} & \textbf{Pixel Accuracy} & \textbf{Mean IOU} & \textbf{FW IOU} \\ \hline
    Our weakly-supervised, 256x256 &  25 & 16.45\%  & 39.66\%  & 9.59\% & 28.12\%  \\ \hline
    Our fully-supervised, 256x256 & 100 & 17.13\%  & 48.34\%  & 12.59\% &29.71\%  \\\hline
    \end{tabular}
    }}
    \caption{Evaluation of semantic segmentation results. 
    }
    \label{tab:seg_lostgan} 
\end{table*}

\begin{table*}[!ht]
    \centering
    \resizebox{0.9\textwidth}{!}{
    \begin{tabular}{c|c|c|c|c|c|c|c|c|c|c|c|c}
        \hline
         mIOU & person & giraffe & motorcycle & zebra & sky & grass & sea & snow & stop sign & tree & clouds & pizza   \\ \hline
         Weakly-supervised & 43.0 & 36.0 & 28.1 & 49.4 & 54.8 & 54.3 & 54.3 & 52.8 & 39.0 & 49.4 & 40.6 & 37.3  \\ \hline
          Fully-supervised & 50.9 & 45.3 & 33.1 & 59.6 & 63.8 & 56.8 & 65.9 & 63.8 & 58.0 & 58.1 & 39.7 & 37.4 \\ \hline
    \end{tabular}
    }
    \caption{Category breakdown of the Mean IOU (mIOU) comparisons between predicted masks and ground truth masks on some selected categories.}
    \label{tab:selected_miou}
\end{table*}

\subsection{Weakly-supervised image semantic segmentation results.}
\textbf{Baselines.} Most of weakly-supervised semantic segmentation methods have not been evaluated in the COCO-Stuff dataset. In the meanwhile, the inference network architecture in our DCL is not directly comparable to those in existing methods. To show the effectiveness of the weakly-supervised inference network in our DCL, we train the inference network individually under the fully-supervised setting as the baseline. 

\textbf{Evaluation Metrics.} We follow the evaluation protocol used in the COCO-Stuff benchmark~\cite{caesar2018coco} including the class accuracy, pixel accuracy, mean IOU and FW (category frequency weighted) IOU. 

\textbf{Results.} Table~\ref{tab:seg_lostgan} shows the quantitative comparisons. The weakly-supervised inference network in our DCL obtains comparable performance with the fully-supervised counterpart, especially in terms of FW IOU. \textit{In terms of the number of epochs, the inference network in our DCL uses much less epochs than the fully-supervised counterpart. This means that the LostGAN is, as a learnable objective function for the inference network, effective.}  Fig.~\ref{fig:teaser} and Fig.~\ref{fig:segmentation_evaluation} show some examples from which we also can see that the proposed DCL is effective in learning the weakly-supervised semantic segmentation. Table~\ref{tab:selected_miou} shows the mean IOU breakdown on some selected categories. \textbf{These results strongly show that our DCL is capable of effectively exploiting generative learning to guide weakly-supervised discriminative learning. } 

\textbf{Remarks.} There are several aspects that can be further studied in our DCL: First, for semantic segmentation, the feature backbone network is typically pretrained on the ImageNet-1k~\cite{deng2009imagenet} dataset. The inference network in our DCL is trained from scratch. Second, the inference network in our DCL uses a very light-weight U-Net architecture without any feature normalization modules (e.g., BatchNorm~\cite{ioffe2015batch}). Removing feature normalization modules in the discriminator network is commonly done in GANs due to the discrepancy of statistics between real and fake images in training, which is applicable to the inference network in our DCL. 
Exploring state-of-the-art neural architectures of image semantic segmentation in the DCL and bringing back feature normalization modules (e.g., BatchNorm) are of great interest in our future work. For the former, we need to study more sophisticated training methods including hyperparameter and optimizer settings. For the latter, we may need to utilize two feature normalization branches for real images and fake images respectively.

\section{Conclusions}
This paper presents a deep consensus learning (DCL) method for joint layout-to-image synthesis and weakly-supervised image semantic segmentation. The proposed DCL is formulated under a three-player minmax game setting, which is built on the LostGANs~\cite{LostGAN,sun2020learning} and introduces an inference network as the third player. The inference network realizes weakly-supervised image semantic segmentation.  The three networks are trained end-to-end using two deep consensus mappings, together with the adversary loss. In experiments, the proposed DCL is tested in the COCO-Stuff dataset \cite{caesar2018coco} at the resolution of 256$\times$256. It obtains compelling layout-to-image synthesis results and weakly-supervised image semantic segmentation results. The proposed DCL sheds light on addressing the problem of simultaneously learning structured input synthesis and weakly-supervised structured output prediction from scratch end-to-end. 

	\section*{Acknowledgements}
	This work was supported in part by NSF IIS-1909644, ARO Grant W911NF1810295, NSF IIS-1822477, NSF CMMI-2024688, NSF IUSE-2013451 and DHHS-ACL Grant 90IFDV0017-01-00. The views presented in this paper are those of the authors and should not be interpreted as representing any funding agencies.

{\small
\bibliographystyle{ieee_fullname}
\bibliography{egbib}
}

\end{document}